\pdfoutput=1

\documentclass[11pt]{article}

\usepackage{EMNLP2022}
\usepackage{graphicx} 

\usepackage{times}
\usepackage{amsmath}
\usepackage{amssymb}
\usepackage{amsthm, bm}
\usepackage{booktabs}
\usepackage{latexsym}
\usepackage{url}
\usepackage{makecell}
\usepackage{multirow}
\usepackage[T1]{fontenc}

\usepackage[utf8]{inputenc}

\usepackage{microtype}

\usepackage{inconsolata}

\title{Revisiting and Advancing Chinese Natural Language Understanding with Accelerated Heterogeneous Knowledge Pre-training}

\author{Taolin Zhang$^{1,2}$, Junwei Dong$^{2,3}$, Jianing Wang$^{1,2}$, Chengyu Wang$^{2}$\thanks{\ \ \ Corresponding author.}, Ang Wang$^{2}$, \\ \bf{Yinghui Liu$^{2}$, Jun Huang$^{2}$, Yong Li$^{2}$, Xiaofeng He$^{1}$}\\
  $^{1}$ East China Normal University, Shanghai, China \\
  $^{2}$ Alibaba Group, Hangzhou, China \\
  $^{3}$ Chongqing University, Chongqing, China \\
  \texttt{zhangtl0519@gmail.com},
  \texttt{chengyu.wcy@alibaba-inc.com}}
\begin{document}
\maketitle
\begin{abstract}
Recently, knowledge-enhanced pre-trained language models (KEPLMs) improve context-aware representations via learning from structured relations in knowledge graphs, and/or linguistic knowledge from syntactic or dependency analysis. Unlike English, there is a lack of high-performing open-source Chinese KEPLMs in the natural language processing (NLP) community to support various language understanding applications.
In this paper, we revisit and advance the development of Chinese natural language understanding with a series of novel Chinese KEPLMs released in various parameter sizes, namely CKBERT (Chinese knowledge-enhanced BERT).
Specifically, both relational and linguistic knowledge is effectively injected into CKBERT based on two novel pre-training tasks, i.e., linguistic-aware masked language modeling and contrastive multi-hop relation modeling.
Based on the above two pre-training paradigms and our in-house implemented TorchAccelerator, we have pre-trained base (110M), large (345M) and huge (1.3B) versions of CKBERT efficiently on GPU clusters.
Experiments demonstrate that CKBERT outperforms strong baselines for Chinese over various benchmark NLP tasks and in terms of different model sizes.
\footnote{All the codes and model checkpoints have been released to public in the EasyNLP framework~\cite{DBLP:journals/corr/abs-2205-00258}. URL:~\url{https://github.com/alibaba/EasyNLP}.}
\end{abstract}

\section{Introduction}
Pre-trained Language Models (PLMs) such as BERT \citep{DBLP:conf/naacl/DevlinCLT19} are pre-trained by self-supervised learning on large-scale text corpora to capture the rich semantic knowledge of words \citep{DBLP:conf/acl/LiZT0R20,DBLP:conf/acl/GongZZSWW22}, improving various downstream NLP tasks significantly \citep{DBLP:conf/emnlp/HeZZCC20, DBLP:conf/acl/XuGTSSGZQJD20, DBLP:conf/acl/ChangXXT20}.
Although these PLMs have stored much internal knowledge \citep{DBLP:conf/emnlp/PetroniRRLBWM19,DBLP:conf/akbc/PetroniLPRWM020}, they can hardly understand external background knowledge from the world such as factual and linguistic knowledge \cite{DBLP:journals/corr/abs-2101-12294, DBLP:conf/acl/CuiCWZ21,DBLP:conf/naacl/LaiLFHZ21}.



In the literature, most approaches of knowledge injection can be divided into two categories, including relational knowledge and linguistic knowledge.
(1) Relational knowledge-based approaches inject entity and relation representations in Knowledge Graphs (KGs) trained by knowledge embedding algorithms \citep{DBLP:conf/acl/ZhangHLJSL19,DBLP:conf/emnlp/PetersNLSJSS19} or convert triples into sentences for joint pre-training \citep{DBLP:conf/aaai/LiuZ0WJD020, DBLP:conf/coling/SunSQGHHZ20}.
(2) Linguistic knowledge-based approaches extract semantic units from pre-training sentences such as part-of-speech tags, constituent and dependency syntactic parsing, and feed all linguistic information into various transformer-based architectures \citep{DBLP:conf/emnlp/Zhou0ZZ20, DBLP:conf/naacl/LaiLFHZ21}.
We observe that there can be three potential drawbacks.
(1) These approaches generally utilize a single source of knowledge (i.e., inherent linguistic knowledge), which ignore important knowledge from other sources \citep{DBLP:journals/aiopen/SuHZLLLZS21} (i.e., relational knowledge from KGs).
(2) Training large-scale KEPLMs from scratch requires high-memory computing devices and is time-consuming, which brings significant computational burdens for users \citep{DBLP:journals/aiopen/ZhangHZKGYQSJGQ21, DBLP:conf/aaai/Zhang0HQTH022}.
(3) Most of these models are pre-trained in English only. There is a lack of powerful KEPLMs for understanding other languages \citep{DBLP:conf/icpr/LeeYHJMG20,DBLP:journals/corr/abs-2111-09453}.

To overcome the above problems, we release a series of Chinese KEPLMs named CKBERT (Chinese knowledge-enhanced BERT), with heterogeneous knowledge sources injected.
We particularly focus on Chinese as it is one of the most widely spoken languages other than English.
The CKBERT models are pre-trained by two well-designed pre-training tasks as follows:
\begin{itemize}
    \item \textbf{Linguistic-aware Masked Language Modeling (LMLM):} 
    LMLM is substantially extended from Masked Language Modeling (MLM) \citep{DBLP:conf/naacl/DevlinCLT19} by introducing two key linguistics tokens derived from dependency syntactic parsing and semantic role labeling. We also insert unique markers for each linguistic component among contiguous tokens. The goal of LMLM is to predict both randomly selected tokens and linguistic tokens masked in the pre-training sentences.

    \item \textbf{Contrastive Multi-hop Relation Modeling (CMRM):} 
    We sample fine-grained subgraphs from a large-scale Chinese KG by multi-hop relations to compensate for understanding the background knowledge of target entities. Specifically, we construct positive triples for matched target entities via retrieving one-hop entities in the corresponding subgraphs. Negative triples are sampled from unrelated multi-hop entities through the relation paths in the KG. 
    The CMRM task is proposed to pull the semantics of similar entities close and push away those with irrelevant semantics.
\end{itemize}

Based on the above heterogeneous knowledge pre-training tasks, we produce various sizes of CKBERT models to meet the inference time and accuracy requirements of different real-world scenarios \citep{DBLP:conf/nips/BrownMRSKDNSSAA20,DBLP:journals/corr/abs-2204-02311}, including base (110M), large (345M) and huge (1.3B).
The models are pre-trained using our in-house implemented TorchAccelerator that effectively transforms PyTorch eager execution to graph execution on distributed GPU clusters, boosting the training speed by 40\% per sample with our advanced compiler technique based on Accelerated Linear Algebra (XLA).
In the experiments, we compare CKBERT against strong baseline PLMs and KEPLMs
on various Chinese general and knowledge-related NLP tasks. The results demonstrate the improvement of CKBERT compared to SoTA models. 
    


\begin{figure*}[t]
\centering
\includegraphics[height=8.75cm]{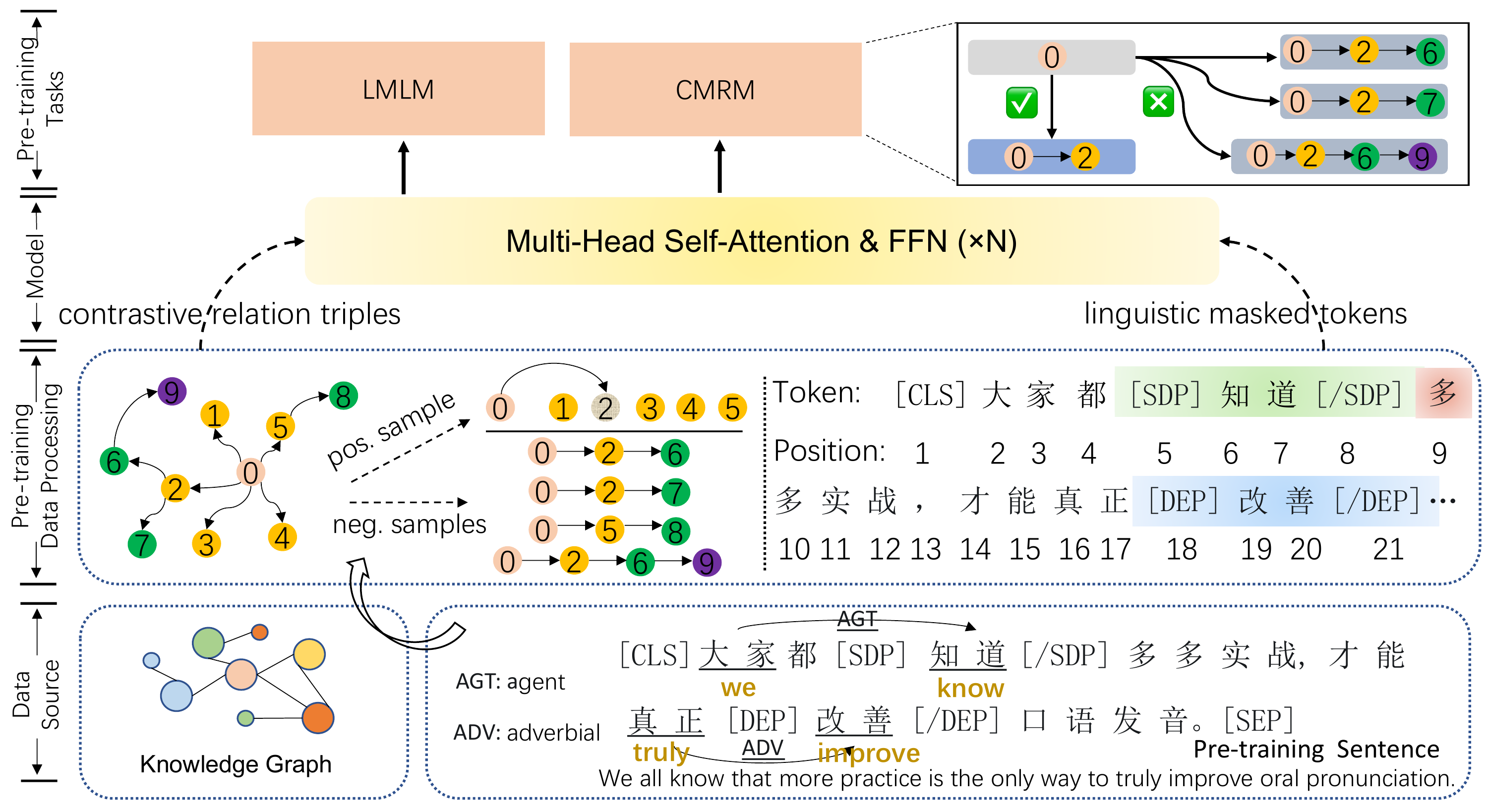}
\caption{Model overview. The LMLM task is not only able to perform random masked token prediction (similar to BERT) but also to predict masked linguistic-aware tokens. The CMRM task injects external relation triples into  PLMs through neighboring multi-hop relations. (Best viewed in color.)}
\label{model_figure}
\end{figure*}

\section{Related Work}
We briefly summarize the related work on the following two aspects: PLMs and KEPLMs.

\subsection{PLMs}
Following BERT~\cite{DBLP:conf/naacl/DevlinCLT19}, 
many PLMs have been proposed to improve performance in various NLP tasks.
Several approaches extend BERT by employing novel token-level and sentence-level pre-training tasks.
Notable PLMs include ERNIE-Baidu \citep{DBLP:journals/corr/abs-1904-09223}, MacBERT \citep{DBLP:conf/emnlp/CuiC000H20} and PERT \citep{DBLP:journals/corr/abs-2203-06906} for Chinese NLU downstream tasks.
Other models boost the performance by changing the internal encoder architectures.
For example, XLNet \citep{DBLP:conf/nips/YangDYCSL19} utilizes Transformer-XL \citep{DBLP:conf/acl/DaiYYCLS19} to encode long sequences by the permutation in language tokens.
Sparse self-attention \citep{DBLP:conf/emnlp/CuiLCZ19} replaces the self-attention mechanism with more interpretable attention units.
Yet, other PLMs such as MT-DNN \cite{DBLP:conf/acl/LiuHCG19} combine self-supervised pre-training with the multi-task supervised learning to improve the performance of various GLUE tasks \cite{DBLP:conf/iclr/WangSMHLB19}.

\subsection{KEPLMs}
These models use structured knowledge or linguistic semantics to enhance the language understanding abilities of PLMs.
We summarize recent KEPLMs grouped into the following four types:
(1) Knowledge-enhancement by linguistic semantics. These works use the linguistic information already available in the pre-training sentences to enhance the understanding ability of PLMs.
Lattice-BERT \citep{DBLP:conf/naacl/LaiLFHZ21} pre-trains a Chinese PLM over a word lattice \citep{DBLP:journals/tacl/BuckmanN18} structure to exploit multi-granularity inputs.
(2) Knowledge-enhancement by entity embeddings. For example, ERNIE-THU \citep{DBLP:conf/acl/ZhangHLJSL19} injects entity embeddings into contextual representations via knowledge-encoders stacked by the information fusion module. 
(3) Knowledge-enhancement by entity descriptions. These approaches learn entity embeddings by knowledge descriptions.
For example, pre-training corpora and entity descriptions in KEPLER \citep{DBLP:journals/tacl/WangGZZLLT21} are encoded into a unified semantic space within the same PLM.
(4) Knowledge-enhancement by converted triplet's texts. K-BERT \citep{DBLP:conf/aaai/LiuZ0WJD020} and CoLAKE \citep{DBLP:conf/coling/SunSQGHHZ20} convert relation triplets into texts and insert them into training samples without using pre-trained embeddings.
In this paper, we argue that aggregating heterogeneous knowledge information can further benefit the context-aware representations of PLMs.

\section{Model}
In this section, we elaborate the techniques of the proposed CKBERT model.
The main architecture of CKBERT is firstly presented in Figure \ref{model_figure}.




\subsection{Model Architecture}
It accepts a sequence of $M$ WordPiece tokens \citep{ DBLP:journals/corr/WuSCLNMKCGMKSJL16}, $\left(x_1, x_2, ..., x_M\right)$ as input, and computes the $D$-dimensional contextual representations $H_i \in \mathbb{R}^{M \times D}$ by successively stacking $N$ transformer encoder layers.
We do not modify the architecture here to guarantee that CKBERT can be seamlessly integrated into any industrial applications that BERT supports with better performance.\footnote{Without loss of generality, we focus on the transformer encoder architecture only; yet our work can also be extended model architectures with slight modification.}

\subsection{Linguistic-aware Masked Language Modeling (LMLM)}

In BERT pre-training, 15\% of all token positions are randomly masked  for prediction. 
However, random masked tokens may be unimportant units such as conjunctions and prepositions \citep{DBLP:conf/blackboxnlp/ClarkKLM19,DBLP:conf/aaai/Hao0W021}.
We reconstruct the input sentences and mask more tokens based on linguistic knowledge 
so that CKBERT can better understand the semantics of important tokens in pre-training sentences.
Specifically, we use the following three steps to mask the linguistic input units:
\begin{itemize}
    \item \textbf{Recognizing Linguistic Tokens:} We first use the off-the-shelf tool\footnote{\url{http://ltp.ai/}} to recognize important units in pre-training sentences, including dependence grammar and semantic dependency parsing.
    The extracted relations here serve as important sources of linguistic knowledge, 
    including ``subject-verb'', ``verb-object'' and ``adverbial'' for dependence grammar and ``non-agent'' for semantic dependency parsing.
    
    \item \textbf{Reconstructing Input Sentences:} In addition to the original input form, based on the subjects and objects of the extracted linguistic relations, we insert special identifiers for each lexicon unit between words spans to give explicit boundary information for model pre-training.
    For example, we add \texttt{[DEP]} and \texttt{[/DEP]} for dependence grammar and \texttt{[SDP]} and \texttt{[/SDP]} for dependency parsing tokens.
    
    \item \textbf{Choosing Masked Tokens:} We choose 15\% of token positions from the reconstructed input sentence for masking, using the special token \texttt{[MASK]}.
    Among these tokens, we assign 40\% of the positions to randomly selected tokens and the rest to linguistic tokens.
    Note that these special identifiers (\texttt{[DEP]}, \texttt{[/DEP]}, \texttt{[SDP]} and \texttt{[/SDP]}) are also treated as normal tokens for masking,  thus the model needs to be aware of predicting word boundaries rather than simply filling in masks based on contexts.
\end{itemize}


After input sentences are processed, for LMLM,
let $\Omega=(m_1, m_2, m_3,...,\gamma_{K-1}, \gamma_K)$ denote the indexes of the masked tokens in the sentence $X$, where $m_i$ is an index of a randomly masked token, $\gamma_i$ is an index of a selected linguistic-aware masked token and $K$ is the total number of masked tokens.
Let $X_\Omega$ denote the set of masked tokens in $X$, and $X_{-\Omega}$ denote the set of observed (unmasked) tokens.
The objective of LMLM is as follows:
\begin{gather}
\label{mlm_equ}
\mathcal{L}_{mlm}(X_\Omega|X_{-\Omega})=\frac{1}{K} \sum_{k=1}^{K}\log p(x_{m_k|\gamma_k}|X_{-\Omega};\theta)
\end{gather}
where $x_{m_k|\gamma_k}$ denotes the randomly selected tokens or the linguistic tokens.
$\theta$ represents the parameter collection of our model.

\subsubsection{Contrastive Multi-hop Relation Modeling (CMRM)}

In addition to LMLM, we further inject relation triples into CKBERT to make it understand the background factual knowledge of entities.
For an entity in the pre-training sentence, we construct positive and negative relation triples as follows:
\begin{itemize}
    \item \textbf{Positive Triples:} We employ entity linking
    to link an entity in the pre-training sentence to the target entity $e_t$ in the KG.
    The relation triples w.r.t. the one-hop entities are viewed as candidate positive triples.
    Next, we choose a relation triple randomly from the candidates as a positive sample, denoted as $t_p$.
    
    \item \textbf{Negative Triples:} Because the semantic similarity between the positive triple $t_p$ and the relation triples along the KG paths decreases, we construct $L$ candidate negative triples ($t_n^1, t_n^2,...,t_n^L$) by making multiple hops starting from the target entity $e_t$.
    For example, in Fig. \ref{model_figure}, we take the target entity $e_0$ as the starting node and retrieve the nodes along the edges. We obtain the ending node $e_{end}$ with multi-hop relations $Hop(\mathcal{G}, e_0, e_{end}, r)$, where $Hop(\cdot)$ means the shortest distance between $e_0$ and $e_{end}$ in KG $\mathcal{G}$.
    Here, we regard a triple to be negative $t_n$ if $Hop(\cdot)>1$ and is no larger than a small threshold $\delta$\footnote{If the threshold for the number of hops $\delta$ is too large, the model can easily distinguish the positive and negative triples due to the large semantic gaps. For effective contrastive learning, good negative triples should be ``hard negatives''.}.
    In this paper, we set $\delta=3$.
    Hence, there are four negative triples for $e_0$ in Figure \ref{model_figure}. A sample three-hop path is $e_0 \to e_2 \to e_6 \to e_9$.
\end{itemize}

\begin{table*}[t]
\centering
\begin{small}
\begin{tabular}{c|cccccc|ccc|c}
\toprule
 \multirow{2}{*}{\bf Model} & \multicolumn{6}{c|}{\bf Text Classification} & \multicolumn{3}{c|}{\bf Question Answering} & {\bf Total}\\
\cmidrule(r){2-7} \cmidrule(r){8-10} 
 & AFQMC & TNEWS  & IFLYTEK & OCNLI & WSC & CSL & CMRC & CHID & C3  & {\bf Score}\\
\midrule
BERT & 72.73 & 55.22 & 59.54 & 66.53  & 72.49  & 81.77 & 73.40 & 79.19 & 57.91 & 69.72  \\
MacBERT & 69.90 & \underline{\textbf{57.93}}  & 60.35 & 67.43 & 74.71  & 82.13 & 73.55 & 79.51 &\underline{\textbf{ 58.89}}  & 70.28 \\
PERT & \underline{\textbf{73.61}} & 54.50 & 57.42 & 66.70 & 76.07 & \underline{\textbf{82.77}} & \underline{\textbf{73.80}} & 80.19 & 58.03 & 70.18 \\\midrule
ERNIE-Baidu & 73.08& 	56.22& 	60.11& 	67.48& 	75.79& 	82.14& 	72.86& 	80.03& 	57.63 & 69.83 \\
Lattice-BERT & 72.96& 	56.14& 	58.97& 	67.54& 	76.10& 	81.99& 	73.47& 	80.24& 	57.80 & 70.29 \\
K-BERT & 73.15& 	55.91& 	60.19& 	67.83& 	76.21& 	82.24& 	72.74& 	80.29& 	57.48 & 70.35\\
ERNIE-THU & 72.88& 	56.59 & 	59.33& 	67.95& 	75.82& 	82.35& 	72.96& 	80.22& 	56.30 & 69.98\\ \midrule
\bf CKBERT-base & 73.17 & 56.44 & \underline{\textbf{60.65}} & \underline{\textbf{68.53}} & \underline{\textbf{76.38}} & 82.63 & 73.55 & \underline{\textbf{81.69}} & 57.91 & \underline{\textbf{71.36}} \\
\midrule
\bf CKBERT-large & \textbf{74.75} & 55.86 & \textbf{60.62} & \textbf{70.57} & \textbf{78.90}  &  82.30 & 73.45 & \textbf{82.34}& 58.12 &\textbf{72.23} \\
\bf CKBERT-huge & \textbf{75.03} & \textbf{59.72} & \textbf{60.96} & \textbf{78.26} & \textbf{85.16}  &  \textbf{89.47} & \textbf{77.25} & \textbf{97.73}& \textbf{86.59} &\textbf{78.91} \\
\bf CKBERT-huge & \multirow{2}{*}{\textbf{77.05}} & \multirow{2}{*}{\textbf{61.16}} & \multirow{2}{*}{\textbf{61.19}} & \multirow{2}{*}{\textbf{82.80}} & \multirow{2}{*}{\textbf{87.14}}  &  \multirow{2}{*}{\textbf{94.23}} & \multirow{2}{*}{\textbf{80.40}} & \multirow{2}{*}{\textbf{97.91}} & \multirow{2}{*}{\textbf{87.26}} & \multirow{2}{*}{\textbf{81.02}} \\
\bf (Ensemble) & & & & & & & & & & \\\bottomrule
\end{tabular}
\end{small}
\caption{Performance of tasks on the CLUE 1.1 testing sets (\%). The ``Total Score'' is the weighted averaged score of the nine tasks generated by the official website automatically. All the baseline models are base models (with the same or similar parameter size as that of BERT-base).
}
\label{general_result}
\end{table*}

The CMRM task is designed for pulling similar relational triples of the target entity closely and pushing unrelated multi-hop relational triples away, in order to enhance the external background knowledge of the target entity from the KG.
Concretely, after the positive sample $t_p$ and negative samples ($t_n^1, t_n^2,...,t_n^L$) of the target entity $e_t$ are retrieved,
the context-aware representations of the target entity $e_t$ can be obtained as follows:
\begin{gather}
h_{e_t}=\mathcal{LN}\left(\sigma\left( f_{sp} \left(h_{e_t^i}, \ldots, h_{e_t^j}\right)\mathrm{W_{1}} \right)\right)
\end{gather}
where $h_{e_t}$ is the hidden representation of the target entity $e_t$ constructed by the entity's token representations $\left(h_{e_t^i},\ldots,h_{e_t^j}\right)$, as an entity can have multiple tokens in the pre-training sentence.
$f_{sp}$ is the self-attentive pooling operator \citep{DBLP:conf/iclr/LinFSYXZB17}, $\sigma(\cdot)$ non-linear activation function GELU \citep{hendrycks2016gaussian} and
$\mathcal{LN}(\cdot)$ is the LayerNorm function \citep{DBLP:journals/corr/BaKH16}.
$\mathrm{W_{1}}$ is the learnable weight matrix.

Meanwhile, as relation triples can be viewed as natural sentences via concatenating the triple's tokens together, following \citet{DBLP:conf/aaai/LiuZ0WJD020,DBLP:conf/coling/SunSQGHHZ20}, we convert the triples into sentences to generate the representations obtained by the shared encoder $\theta$ (which is the transformer encoder of our CKBERT model).
Hence, the representations of the positive triple $h_{t_p}$ and the negative triples ($h_{t_n^1}, h_{t_n^2},...,h_{t_n^L}$) can also be derived.
For the CMRM task, we employ InfoNCE \citep{DBLP:journals/corr/abs-1807-03748} as the loss function to calculate the similarity as follows:
\begin{equation}
\mathcal{L}_{cl}=-\log \frac{\exp \left(cos \left(h_{e_{t}}, h_{t_p}\right) / \tau\right)}{\sum_{l=1}^{L} \exp \left(cos \left(h_{e_{t}},h_{t_n^l} \right) / \tau\right)}
\end{equation}
where $cos(\cdot, \cdot)$ denotes the cosine function to calculate the similarity between entity and relation representations, and $\tau$ is a pre-defined hyper-parameter.

\subsection{Optimization of Model Pre-training}

For model training optimization, we first give the total loss function for pre-training CKBERT based on our two novel pre-training tasks as follows:
\begin{equation}
\mathcal{L}_{total} = \mathcal{L}_{mlm} + \mathcal{L}_{cl}
\end{equation}
Here, we pre-train a series of CKBERT models on distributed GPU clusters, with codes in PyTorch. As PyTorch employs eager execution for tensor computation, it lacks graph-based intermediate representations of models, hindering deeper optimization~\cite{DBLP:conf/nips/PaszkeGMLBCKLGA19}.

Inspired by LazyTensor~\cite{DBLP:journals/corr/abs-2102-13267} and Pytorch/XLA on cloud TPUs\footnote{\url{https://github.com/pytorch/xla}}, we develop the TorchAccelerator toolkit for Pytorch training acceleration on GPU clusters.
Through XLA custom function and code parsing with an abstract syntax tree (AST), we improve the completeness and performance of the transformation from eager execution to graph execution. A computational graph is  generated by TorchAccelerator. The operators on the graph will be fused. By fusing operators, the kernel launch overhead can be reduced. Moreover, fewer intermediate results are written to memory thus reducing the memory bandwidth usage. The effectiveness of computation is also improved by multi-stream optimization and asynchronous transmission of tensors. Since the implementation of TorchAccelerator is not our major focus, more details will be presented in our future work.

\section{Experiments}
We present comprehensive evaluation results of CKBERT.
Due to space limitation, the details of data sources, baselines and hyper-parameter settings are shown in Appendices \ref{dataset_statistics}, \ref{baselines_exp}, \ref{model_para_traing_details}.


\subsection{General Experimental Results}
We evaluate CKBERT over a widely-used Chinese benchmark CLUE  \citep{DBLP:conf/coling/XuHZLCLXSYYTDLS20} and knowledge-intensive tasks to evaluate the influence of knowledge injection in CKBERT.

\subsubsection{Results of CLUE Benchmark}
The CLUE benchmark contains nine text classification and question answering tasks.
Specifically, the text classification tasks contain various text task types, including the classification of short sentences and long sentence pairs.
The results of all tasks are shown in Table \ref{general_result}.

From the results, we have the following observations. (1) The performance of KEPLMs has a large gap over BERT.
It indicates that the injection of different knowledge sources enables the models to perform better semantic reasoning compared to pre-training on texts only.
(2) The performance of CKBERT is further improved compared to previous strong baseline KEPLMs under the same parameter size in most cases.
From this phenomenon, we believe that the heterogeneous knowledge sources injected into the PLMs benefit the model's results.
(3) The larger the number of parameters in the model, the more effective the heterogeneous knowledge fusion is for downstream tasks.
The huge model of CKBERT (1.3B parameters) outperforms base (110M) by a large margin, which is suitable for applications that require high prediction accuracy.
We also build an ensemble of the huge models (denoted as CKBERT-huge (Ensemble)) from different checkpoints. The performance can be further improved by more than 2.0\%.

\begin{table}[t]
\centering
\begin{footnotesize}
\begin{tabular}{c|cccc}
\toprule
\textbf{Model} & \textbf{MSRA} & \textbf{Weibo} & \textbf{Onto.} & \textbf{Resu.} \\ \midrule
BERT & 95.20 & 54.65 & 81.61 & 94.86  \\
MacBERT & 95.07 & 54.93 & 81.96 & 95.22 \\ 
PERT & 94.99 & 53.74 & 81.44 & 95.10 \\ \midrule
ERNIE-BD & \underline{\textbf{95.39}} & 55.14 & 81.17 & 95.13 \\ 
Lat.-BERT & 95.28 & 54.99 & 82.01 & 95.31 \\ 
K-BERT & 94.97 & 55.21 & 81.98 & 94.92 \\ 
ERNIE-THU & 95.25 & 53.85  & 82.03 & 94.89 \\ \midrule
\bf CKBERT-base & 95.35  & \underline{\textbf{55.97}} & \underline{\textbf{82.19}} & \underline{\textbf{95.68}} \\
\midrule
\bf CKBERT-large & \textbf{95.98}  & \textbf{57.09} & \textbf{82.43} & \textbf{96.08} \\
\bf CKBERT-huge & \textbf{96.79}  & \textbf{58.66} & \textbf{83.87} & \textbf{97.19} \\\bottomrule
\end{tabular}
\end{footnotesize}
\caption{\label{ner_task_result} Performance of CKBERT and baselines over four public Chinese NER datasets in term of F1 (\%).}
\end{table}

\subsubsection{Results of NER}
We further evaluate CKBERT over the four public NER datasets, including MSRA \citep{DBLP:conf/acl-sighan/Levow06}, Weibo \citep{DBLP:conf/emnlp/PengD15}, Ontonotes 4.0\footnote{\url{https://catalog.ldc.upenn.edu/LDC2013T19}}, and Resume \citep{DBLP:conf/ranlp/YangZD17}.
The detailed statistics including the split sizes of training, development, and testing sets are described in Appendix \ref{dataset_statistics}.
The models are stacked by the CKBERT encoder and a softmax linear layer, whose parameters are initialized randomly.
The entities recognized in the samples are labeled by the \texttt{B/I/O/S} tags. This transforms the NER task into a 4-class classification task for each token.

Table \ref{ner_task_result} shows the performance of various models on four NER datasets.
It can be seen that KEPLMs outperform vanilla PLMs.
In addition, our CKBERT model (base) with linguistic and external knowledge achieves a large gap performance compared to baselines.
We believe that heterogeneous knowledge sources play an important role as described in the ablation study (See Section \ref{ablation_study}).

\subsection{Ablation Study}
\label{ablation_study}
In this part, we evaluate the effectiveness of two important model components of CKBERT on representative tasks.
Specifically, We introduce several variants of CKBERT removing certain components.
CKBERT-LMLM means that we remove the LMLM task and only learns the CMRM task during pre-training.
CKBERT-CMRM remove the CRMR task and only perform the LMLM task.
We also provide the results of continual pre-training of BERT-base to remove the influence of additional data sources of plain texts.
The performance of those variants and CKBERT on the testing sets of these datasets are shown in Table \ref{ablation_study_table}.

From the results, we can see that (1) Comparing CKBERT-large to BERT-large (continual pre-trained with the same pre-training data),  the explicit heterogeneous knowledge is more useful than the implicit text corpus for various downstream tasks. (2) We also find that the LMLM pre-training task benefits the QA and NER tasks more, whereas the CMRM task improves the performance of plain NLU task (i.e., text classification) significantly.
We conjecture that the main reason behind this phenomenon is that the external background knowledge can easily boost the performance due to the shallow semantics of these simple tasks \citep{DBLP:journals/corr/abs-2110-00269}.

\begin{table}[t]
\centering
\begin{footnotesize}
\begin{tabular}{c|cccc}
\toprule
\bf Model & \multicolumn{1}{c}{\bf AFQ.}  & \multicolumn{1}{c}{\bf IFLY.} & \multicolumn{1}{c}{\bf CMRC} & \multicolumn{1}{c}{\bf Weibo} \\ \midrule
BERT-large-con. & 73.96 & 60.35 & 73.42 & 56.12 \\
CKBERT-large & \textbf{74.75} & \textbf{60.62} & \textbf{73.45} & \textbf{57.09} \\ \midrule
\quad w/o. LMLM & 73.56 & 60.38 & 73.2 & 56.48 \\
\quad w/o. CMRM & 72.96 & 59.38 & 74.8 & 56.72  \\ \bottomrule
\end{tabular}
\end{footnotesize}
\caption{The performance of models for ablation study. ``AFQ.'' and ``IFLY.'' refer to AFQMC and IFLYTEX, respectively (\%).}
\label{ablation_study_table}
\end{table}
\subsection{Results of TorchAccelerator}
\begin{figure}[t]
\centering
\includegraphics[height=7cm]{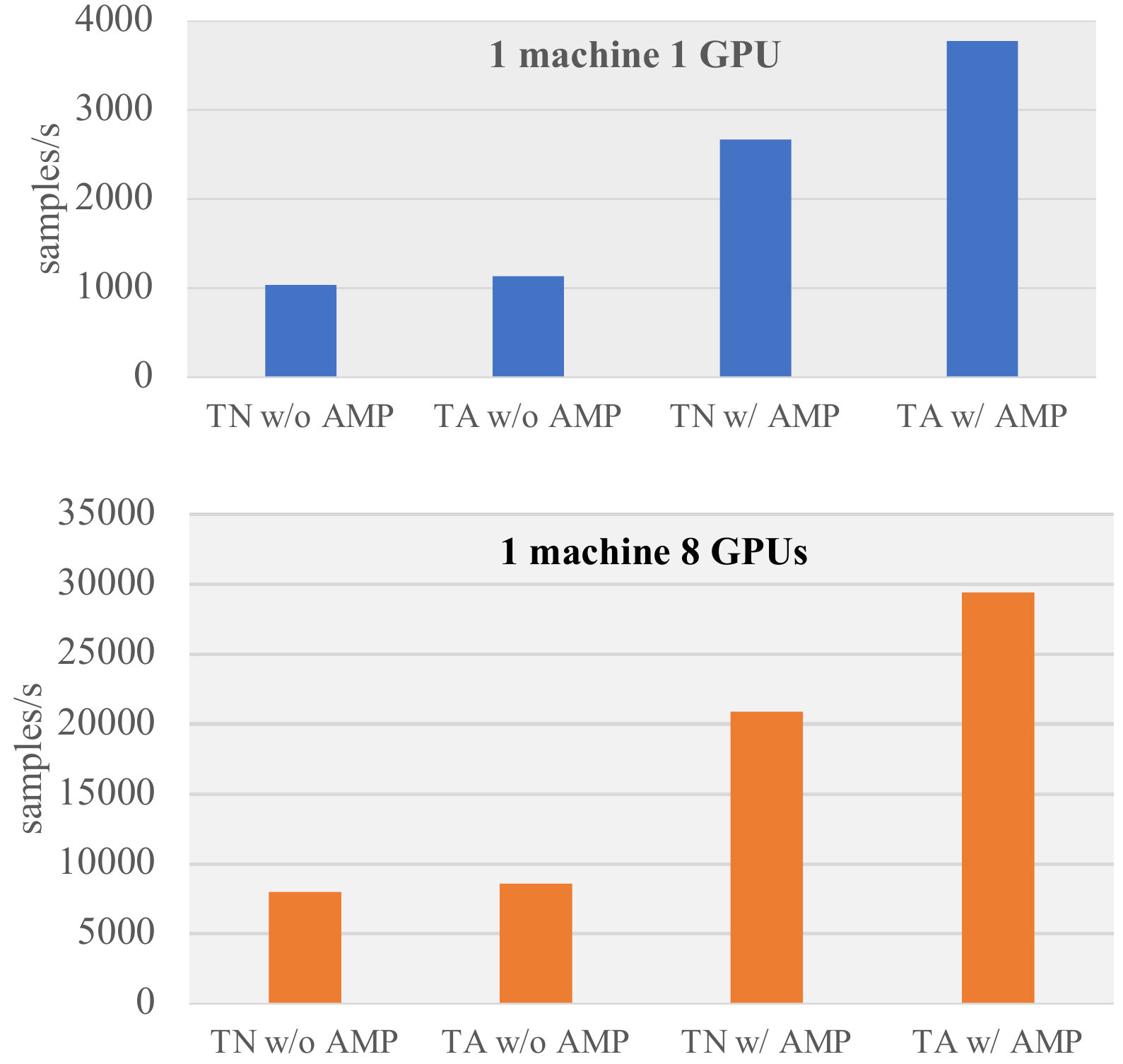}
\caption{Training speed comparison between TorchAccelerator (TA) and Torch Native (TN).
}
\label{torchxla_speed}
\end{figure}

We investigate to what extent  the pre-training speed is improved when our framework is integrated with TorchAccelerator.
Figure~\ref{torchxla_speed} shows the comparison results between TorchAccelerator and Torch Native with AMP (Automatic Mixed Precision) \footnote{\url{https://github.com/NVIDIA/apex}}.
The metric ``samples/s'' means how many samples are computed by the model in each second.
Note that we increase the batch size as large as possible to increase the GPUs' memory utilization and occupancy to 100\%, and thus the experiments w/ and w/o. TorchAccelerator consume the same amount of computational resources. The underlying GPU is Tesla V100 32GB.

From the results, our observations are as follows.
(1) When we only use TorchAccelerator without AMP, the training speed increases slightly.
(2) The training speed can have a large improvement with the interaction between TorchAccelerator and AMP (+40\%).
This is because the kernel fusion of XLA in TorchAccelerator largely reduces the amount of memory access operations, which are the performance bottleneck when AMP is applied.
Hence, our TorchAccelerator effectively reduces the consumption of resources and time during pre-training.

\section{Conclusion and Future Work}
In this paper, we propose a novel series of Chinese KEPLMs named CKBERT to inject the heterogeneous sources including linguistic and external knowledge into the PLMs.
Specifically, we design two novel pre-training tasks including linguistic-aware MLM and contrastive multi-hop relation modeling, and accelerate model pre-training by TorchAccelerator.
The experiments show that our CKBERT outperforms various strong baselines including general PLMs and KEPLMs significantly over knowledge-intensive and natural language understanding tasks.
Future work includes (1) integrating more knowledge sources into PLMs to further improve the performance of downstream tasks; (2) exploring heterogeneous knowledge injection to generative KEPLMs and other languages; and (3) enriching the functionalities of TorchAccelerator and releasing it to public.



\section*{Ethical Considerations}
Our contribution in this work is fully methodological, namely a novel series of KEPLMs, achieving the performance improvement of downstream tasks with different parameter sizes.
Hence, there is no explicit negative social influences in this work.
However, transformer-based models may have some negative impacts, such as gender and social bias. 
Our work would unavoidably suffer from these issues.
We suggest that users should carefully address potential risks 
when the CKBERT models are deployed online.

\section*{Acknowledgments}
\label{sec:acknowledge}
This work has been supported by 
Alibaba Group through Alibaba Innovative Research Program and Alibaba Research Intern Program.

\bibliography{anthology,custom}
\bibliographystyle{acl_natbib}

\begin{table*}[]
\centering
\begin{tabular}{m{3.4cm}<{\centering}  |m{1.5cm}<{\centering} m{1.5cm}<{\centering} m{1.5cm}<{\centering} m{2cm}<{\centering} m{1.5cm}<{\centering}}
\toprule
\bf Dataset & \bf\# Train & \bf\#  Dev & \bf\#  Test & \bf Task & \bf Metric            \\ \midrule
\makecell[c]{AFQMC}  & 34,334 & 4,316 & 3,861 & TC & Acc@1 \\ \midrule
\makecell[c]{TNEWS}  & 53,360 & 10,000 & 10,000 & TC & Acc@1  \\ \midrule
\makecell[c]{IFLYTEK}  & 12,133 & 2,599 & 2,600 & TC & Acc@1  \\ \midrule
\makecell[c]{OCNLI}   & 50,000 & 3,000 & 3,000 & TC & Acc@1  \\ \midrule
\makecell[c]{WSC}  & 1,244 & 304 & 2573 & TC & Acc@1  \\ \midrule
\makecell[c]{CSL}  & 20,000 & 3,000 & 3,000 & TC & Acc@1  \\ \midrule
\makecell[c]{CMRC}  & 10,142 & 1,002 & 3,219 & QA & F1  \\ \midrule
\makecell[c]{CHID}  & 84,709 & 3,218 & 3,231 & QA & Acc@1  \\ \midrule
\makecell[c]{C3}  & 11,869 & 3,816 & 3,892 & QA & Acc@1  \\ \midrule
\makecell[c]{MSRA}  & 40,000 & 6,675 & 4364 & NER & F1  \\ \midrule
\makecell[c]{Resume}  & 3,821 & 462 & 476 & NER & F1  \\ \midrule
\makecell[c]{Weibo}  & 1,350 & 264 & 262 & NER & F1  \\ \midrule
\makecell[c]{Ontonotes}  & 15,740 & 4300 & 4345 & NER & F1  \\ \bottomrule
\end{tabular}
\caption{\label{dataset_statistical_data} The data statistics and evaluation metrics
used in the experiments.}
\end{table*}

\begin{table*}[t]
\centering
\begin{tabular}{c|cccccc}
\toprule
\bf Model & $n_{param.}$ & $N_{layer}$ & $N_{head}$ & $D_{head}$ & $D_{ff}$ & $D_{model}$ \\ \midrule
CKBERT-base & 110M & 12 & 12 & 64 & 3072 & 768 \\
CKBERT-large & 345M & 24 & 16 & 64 & 4096 & 1024 \\ 
CKBERT-huge & 1.3B & 24 & 8 & 256 & 8192 & 2048 \\ 
\bottomrule
\end{tabular}
\caption{The overview of hyper-parameters settings of our model architectures. $n_{param.}$ means the total parameters of our model. $N_{layer}$ is the number of model layers. $N_{head}$ is the number of attention heads in each layer. $D_{head}$ is the hidden dimension of attention heads.
$D_{ff}$ is the intermediate dimension of FFN layers. $D_{model}$ is the output dimension of the model.}
\label{mdoel_architecture}
\end{table*}

\begin{table*}[t]
\centering
\begin{tabular}{c|ccccccc}
\toprule
\bf Dataset & AFQMC & TNEWS & IFLY. & OCNLI & WSC & CSL & CMRC\\ \midrule
BS & 4 & 12 & 4 & 32 & 32 & 16 & 16 \\ \midrule
Epoch & 10 & 10 & 10 & 50 & 50 & 10 & 10\\ \midrule 
LR & 5e-5 & 5e-5 & 5e-5 & 5e-5 & 5e-5 & 5e-5 & 5e-5\\ \midrule
MSL & 256 & 128 & 256 & 128 & 128 & 256 & 512\\
\midrule
\bf Dataset & CHID & C3 & MSRA & Resume & Weibo & Ontonotes      \\ \midrule
BS & 4 & 4 & 32 & 32 & 32 & 32 \\ \midrule
Epoch & 15 & 15 & 10 & 10 & 10 & 10 \\ \midrule 
LR & 5e-5 & 5e-5 & 5e-5 & 5e-5 & 5e-5 & 5e-5 \\ \midrule
MSL & 192 & 512 & 128 & 128 & 128 & 128   \\ \bottomrule
\end{tabular}
\caption{\label{hyperparameters_settings} The important fine-tuning hyper-parameters used in our CKBERT models. ``BS'', ``LR'', and ``MSL'' indicate the batch size, the learning rate and the max sequence length, respectively.}
\end{table*}

\appendix

\section{Data Statistics}
\label{dataset_statistics}
\subsection{Data Sources for Pre-training}
The pre-training corpora after pre-processing contain 5 million text segments with 623,366,851 tokens (6.2 GB).
We also perform simple data pre-processing on the these corpora to improve the quality of the data, including removing incorrect characters and non-Chinese characters, etc.
Our KG data is downloaded from the largest authoritative Chinese KG website OpenKG \footnote{\url{http://openkg.cn/}}.
The number of entities and triples of OpenKG are 16,474,936 and 140,883,574, respectively.
The total number of relation types is 480,882.

\subsection{Statistics of Downstream Tasks}
In this paper, we choose three types downstream tasks for evaluation, including Text Classification (TC), Question Answering (QA) and Named Entity Recognition (NER).
The statistics of dataset sizes are shown in Table \ref{dataset_statistical_data}.
The result metrics used in our models are different among tasks. We use the Acc@1 for text classification, F1 for NER.
For QA tasks, since CHID and C3 tasks are multiple choices, we use Acc@1 as the metric for the two tasks and F1 for CMRC.

\section{Baselines}
\label{baselines_exp}
In this work, we compare CKBERT with general PLMs and KEPLMs with knowledge triples injected, pre-trained on our text corpora:

\subsection{General PLMs}

We use three strong Chinese BERT-style models as baselines, namely BERT-base \citep{DBLP:conf/naacl/DevlinCLT19}, MacBERT \citep{DBLP:conf/emnlp/CuiC000H20} and PERT \citep{DBLP:journals/corr/abs-2203-06906}.
All the model weights are initialized from \citet{DBLP:conf/emnlp/CuiC000H20}.

\subsection{KEPLMs}

We employ three SoTA KEPLMs continually pre-trained on our pre-training corpora as our baseline models, including ERNIE-Baidu~\citep{DBLP:journals/corr/abs-1904-09223}, ERNIE-THU \citep{DBLP:conf/acl/ZhangHLJSL19} and K-BERT \citep{DBLP:conf/aaai/LiuZ0WJD020}.
For a fair comparison, KEPLMs using other resources rather than the KG triples are excluded in this work. All the baseline KEPLMs are injected by the same KG triples during pre-training.

\section{Hyper-parameters Settings}
\label{model_para_traing_details}
\subsection{Hyper-parameters of Pre-training}
For optimization, we set the learning rate as 5e-5, the max sequence length as 128, and the batch size as 20.
The hidden dimension of the text encoder is 768. 
The temperature hyper-parameter $\tau$ is set to 0.5.
The number of negative samples $L$ is $3$.
During pre-training, all the experiments are conducted on 15 servers, each with 8 Tesla V100 GPUs (32GB).

\subsection{Hyper-parameters of Model Architectures}
Table \ref{mdoel_architecture} shows the hyper-parameters settings of our CKBERT models w.r.t. the model architectures, including base (110M), large (345M) and huge (1.3B).

\subsection{Hyper-parameters of Fine-tuning}
Table \ref{hyperparameters_settings} shows the hyper-parameters settings for fine-tuning.
For fair comparison, we set a unified set of important hyper-parameters for each task, including the batch size, the learning epoch, the learning rate and the max sequence length.

\end{document}